%% file: main.tex
\documentclass[10pt,twocolumn,letterpaper]{article}

\usepackage{iccv}
\usepackage{times}
\usepackage{epsfig}
\usepackage{graphicx}
\usepackage{amsmath}
\usepackage{amssymb}
\usepackage{subcaption}
\usepackage{stfloats}
\usepackage[title]{appendix}
\newcommand{\comment}[1]{}
\usepackage{multirow}

\usepackage[pagebackref=true,breaklinks=true,letterpaper=true,colorlinks,bookmarks=false]{hyperref}

\iccvfinalcopy 

\def\iccvPaperID{1779} 

\ificcvfinal\fi
\begin{document}

\title{Shape-Aware Human Pose and Shape Reconstruction Using Multi-View Images} 

\author{Junbang Liang \quad\quad Ming C. Lin\\
University of Maryland, College Park\\
{\tt\small \{liangjb,lin\}@cs.umd.edu}
}

\maketitle
\ificcvfinal\thispagestyle{empty}\fi

\begin{abstract}
We propose a scalable neural network framework to reconstruct the 3D mesh of a human body from multi-view images, in the subspace of the SMPL model~\cite{loper2015smpl}.  Use of multi-view images can significantly reduce the projection ambiguity of the problem, increasing the reconstruction accuracy of the 3D human body under clothing. Our experiments show that this method benefits from the synthetic dataset generated from our pipeline since it has good flexibility of variable control and can provide ground-truth for validation.
Our method outperforms existing methods on real-world images, especially on shape estimations.
\end{abstract}

\input{1-intro}

\input{2-related}
\input{3-statement}

\input{4-network}
\input{5-data}
\input{6-result}

\input{7-conclusion}

{\small
\bibliographystyle{ieee_fullname}
\bibliography{main}
}
\input{appendix}
\end{document}

%% file: 1-intro.tex
\vspace{-1em}
\section{Introduction}
Human body reconstruction, consisting of pose and shape estimation, has been widely studied in a variety of areas, including digital surveillance, computer animation, special effects, and virtual/augmented environments.  Yet, it remains a challenging and popular topic of interest.
While direct 3D body scanning can provide excellent and sufficiently accurate results, its adoption is somewhat limited by the required specialized hardware.
We propose a practical method that can estimate body pose and shape directly from a small set of images (typically 3 to 4) taken at several different view angles, which can be adopted in many applications, such as Virtual Try-On.
Compared to existing scanning-based reconstruction, ours is much easier to use.
Compared to previous image-based estimation methods, ours has a higher shape estimation accuracy when the input human body is not within a normal range of body-mass index (BMI) and/or when the body is wearing loose clothing.
Furthermore, our framework is flexible in the number of images used, which considerably extends its applicability.

In contrast to many existing methods, we use multi-view images as input.
We use the word ``multi-view'' to refer photos taken of the same person with {\em similar} poses from different view angles.  They can be taken using specialized multi-view cameras, but it is not necessary (Sec.~\ref{sec6-3}).
Single-view images often lack the necessary and complete information to infer the pose and shape of a human body, due to the nature of projection transformation.
Although applying a predefined prior can alleviate this ambiguity, it is still insufficient in several cases, especially when a part of the body is occluded by clothing, or when the pose direction is perpendicular to the camera viewing plane.
For example, when the human is walking towards the camera, it can be difficult to distinguish the difference between a standing vs. walking pose using a direct front-view image, while a side-view image could be more informative of the posture.
By obtaining information from multiple view angles, the ambiguity from projection can be considerably reduced, and the body shape under loose garments can also be more accurately recovered.

Previous work on pose and shape estimation of a human body (see Sec.~\ref{sec2}) mostly rely on optimization.
One of the most important metrics used in these methods is the difference between the original and the estimated silhouette.
As a result, these methods cannot be directly applied to images where the human wears loose garments, e.g. long coat, evening gown.
The key insight of our method is: when estimating a person's shape, how the human body is interacting with the cloth, \eg how a t-shirt is stretched out as pushed by the stomach or the chest, provides more information than the silhouette of the person.
So image features, especially those on clothes, play an important role in the shape estimation.
With recent advances in deep learning, it is widely believed that the deep Convolutional Neural Network (CNN) structure can effectively capture these subtle visual details as activation values.
We propose a multi-view multi-stage network structure to effectively capture visual features on garments from different view angles to more accurately infer pose and shape information.

Given a limited number of images, we incorporate prior knowledge about the human body shape to be reconstructed.
Specifically, we propose to use the Skinned Multi-Person Linear (SMPL) model~\cite{loper2015smpl}, which uses Principal Component Analysis (PCA) coefficients to represent human body shapes and poses.
In order to train the model to accurately output the coefficients for the SMPL model, a sufficient amount of data containing ground-truth information is required.
However, to the best of our knowledge, no such dataset exists to provide multiple views of a loosely clothed body with its ground-truth shape parameters (i.e. raw mesh).
Previous learning-based methods do not address the shape (geometry) recovery problem~\cite{mehta2017vnect} or only output one approximation close to the standard mean shape of the human body~\cite{hmrKanazawa17}, which is insufficient when recovering human bodies with largely varying shapes.
Taking advantage of physically-based simulation, we design a system pipeline to generate a large number of multi-view human motion sequences with different poses, shapes, and clothes.
By training on the synthetic dataset with ground-truth shape data, our model is ``shape-aware'', as it captures the statistical correlation between visual features of garments and human body shapes.
We demonstrate in the experiments that the neural network trained using additional simulation data can considerably enhance the accuracy of shape recovery.

To sum up, the key contributions of our work include:
\begin{itemize}
\vspace{-0.5em}
    \item A learning-based {\em shape-aware} human body mesh reconstruction using SMPL parameters for both pose and shape estimation that is supervised directly on shape parameters.
\vspace{-0.75em}
\item A scalable, end-to-end, multi-view multi-stage learning framework to account for the ambiguity of the 3D human body (geometry) reconstruction problem from 2D images, achieving improved estimation results.
\vspace{-0.75em}
\item A large simulated dataset, including {\em clothed} human bodies and the corresponding ground-truth parameters, to enhance the reconstruction accuracy, especially in shape estimation, where no ground-truth or supervision is provided in the real-world dataset.
\vspace{-0.75em}
    \item Accurate {\em shape} recovery {\em under occlusion of garments} by (a) providing the corresponding supervision and (b) deepening the model using the multi-view framework.
\end{itemize}

%% file: 2-related.tex
\vspace{-1em}
\section{Related Work}
\label{sec2}
In this section, we survey recent works on human body pose and shape estimation, neural network techniques, and other related work that make use of synthetic data.

\vspace{-0em}
\subsection{Human Body Pose and Shape Recovering}
Human body recovery has gained substantial interest due to its importance in a large variety of applications, such as virtual environments, computer animation, and garment modeling.
However, the problem itself is naturally ambiguous, given limited input and occlusion.
Previous works reduce this ambiguity using different assumptions and input data.
They consist of four main categories: pose from images, pose and shape from images under tight clothing, scanned meshes, and images with loose clothing.

\noindent
\textbf{Pose From Images.} Inferring 2D or 3D poses in images of one or more people is a popular topic in Computer Vision and has been extensively studied~\cite{pavlakos2017coarse,tekin2017learning,tome2017lifting,zhou2017weaklysupervised,zhou2016deep}.
We refer to a recent work, VNect by Mehta \etal~\cite{mehta2017vnect} that is able to identify human 3D poses from RGB images in real time using a CNN.
By comparison, our method estimates the pose and shape parameters at the same time, recovering the entire human body mesh rather than only the skeleton.

\noindent
\textbf{Pose and Shape From Images under Tight Clothing.} 
Previous work~\cite{balan2007detailed,chen2010inferring,dibra2016hs,guan2009estimating,hasler2010multilinear,jain2010moviereshape} use the silhouette as the main feature or optimization function to recover the shape parameters.
As a result, these methods can only be used when the person is wearing tight clothes, as shown in examples~\cite{tan2017indirect,tung2017self}.
By training on images with humans under various garments both in real and synthetic data, our method can learn to capture the underlying human pose and shape based on image features.

\noindent
\textbf{Pose and Shape From Scanned Meshes.} One major challenge of recovering human body from scanned meshes is to remove the cloth mesh from the scanned human body wearing clothes~\cite{pons2017clothcap}.
Hasler \etal~\cite{hasler2009estimating} used an iterative approach.
They first apply a Laplacian deformation to the initial guess, before regularizing it based on a statistical human model.
Wuhrer \etal~\cite{wuhrer2014estimation} used landmarks of the scanned input throughout the key-frames of the sequences to optimize the body pose, while recovering the shape based on the `interior distance' that helps constrain the mesh to stay under the clothes, with temporal consistency from neighboring frames.
Yang \etal~\cite{yang2016estimation} applies a landmark tracking algorithm to prevent excessive human labor.
Zhang \etal~\cite{zhang2017detailed} took more advantages of the temporal information to detect the skin and cloth region.
As mentioned before, methods based on scanned meshes are limited: the scanning equipment is expensive and not commonly used.
Our method uses RGB images that are more common and thus much more widely applicable.

\noindent
\textbf{Pose and Shape from Images under Clothing.} B\u{a}lan \etal~\cite{bualan2008naked} are the first to explicitly estimate pose and shape from images of clothed humans.
They relaxed the loss on clothed regions and used a simple color-based skin detector as an optimization constraint.
The performance of this method can be easily degraded when the skin detector is not helpful, \eg when people have different skin colors or wear long sleeves.
However, our method is trained on a large number of images, which does not require this constraint.
Bogo \etal~\cite{bogo2016keep} used 2D pose machines to obtain joint positions and optimizes the pose and shape parameters based on joint differences and inter-penetration error.
Lassner \etal~\cite{lassner2017unite} created a semi-automatic annotated dataset by incorporating a silhouette energy term on SMPLify~\cite{bogo2016keep}.
They trained a Decision Forest to regress the parameter based on a much more dense landmark set provided by the SMPL model~\cite{loper2015smpl} during the optimization.
Constraining the silhouette energy effect to a human body parameter subspace can reduce the negative impact from loose clothing, but their annotated data are from the optimization of SMPLify~\cite{bogo2016keep}, which has introduced errors inherently.
In contrast, we generate a large number of human body meshes wearing clothes, with the pose and shape ground-truth, which can then train the neural network to be ``{\em shape-aware}''.

\vspace{-0em}
\subsection{Learning-Based Pose/Shape Estimations}
Recently a number of methods have been proposed to improve the 3D pose estimation with calibrated multi-view input, either using LSTM~\cite{trumble2017total,nunez2019multiview}, auto-encoder~\cite{rhodin2018unsupervised,trumble2018deep} or heat map refinement~\cite{pavlakos2017harvesting,tome2018rethinking}.
They mainly focus on 3D joint positions without parameterization, thus not able to articulate and animate.
Choy \etal~\cite{choy20163d} proposed an LSTM-based shape recovery network for general objects.
Varol \etal~\cite{varol2018bodynet} proposed a 2-step estimation on human pose and shape.
However, both methods are largely limited by the resolution due to the voxel representation.
In contrast, our method outputs the entire body mesh with parameterization, thus is articulated with a high-resolution mesh quality.
Also, our method does not need the calibration of the camera, which is more applicable to in-the-wild images.
Kanazawa \etal~\cite{hmrKanazawa17} used an iterative correction framework and regularized the model using a learned discriminator.
Since they do not employ any supervision other than joint positions, the shape estimation can be inaccurate, especially, when the person is relatively over-weighted.
In contrast, our model is more shape-aware due to the extra supervision from our synthetic dataset.
Recent works~\cite{omran2018neural,pavlakos2018learning,kolotouros2019convolutional} tackle the human body estimation problem using various approaches; our method offers better performance in either single- or multi-view inputs by comparison (see Appendix~\ref{secc}).

\vspace{-0em}
\subsection{Use of Synthetic Dataset}
Since it is often time- and labor-intensive to gather a dataset large enough for training a deep neural network, an increasing amount of attention is drawn to synthetic dataset generation.
Recent studies~\cite{chen2016synthesizing,yang2017learning} have shown that using a synthetic dataset, if sufficiently close to the real-world data, is helpful in training neural networks for real tasks.
Varol \etal~\cite{varol2017learning} built up a dataset (SURREAL) which contains human motion sequences with clothing using the SMPL model and CMU MoCap data~\cite{cmumocap}.
While the SURREAL dataset is large enough and is very close to our needs, it is still insufficient in that (a) the clothing of the human is only a set of texture points on the body mesh, meaning that it is a tight clothing, (b) the body shape is drawn from the CAESAR dataset~\cite{robinette2002civilian}, where the uneven distribution of the shape parameters can serve as a ``prior bias'' to the neural network, and (c) the data only consists of single view images, which is not sufficient for our training.
Different from~\cite{chen2016synthesizing,varol2017learning}, our data generation pipeline is based on physical simulation rather than pasting textures on the human body, enabling the model to learn from more realistic images where the human is wearing looser garments.
Recent works~\cite{sattar2019fashion,alldieck2019learning} also generate synthetic data to assist training, but their datasets have only very limited variance on pose, shape, and textures to prevent from overfitting.
In contrast, our dataset consists of a large variety of different poses, shapes, and clothing textures.

%% file: 3-statement.tex
\vspace{-0.5em}
\section{Overview}
\label{sec3}
In this section, we give an overview of our approach.  First, we define the problem formally.
Then, we introduce the basic idea of our approach.

\noindent
\textbf{Problem Statement: }Given a set of multi-view images, ${\mathbf I}_1$ \ldots ${\mathbf I}_n$, taken for the same person with the same pose, recover the underlying human body pose and shape.

In the training phase, we set $n=4$, \ie by default we take four views of the person: front, back, left and right, although the precise viewing angles and their orders are not required, as shown in Sec.~\ref{sec4-3}.
To extend our framework to be compatible with single view images, we copy the input image four times as the input.
For more detail about image ordering and extensions to other multi-view input, please refer to Sec.~\ref{sec4-3}.
We employ the widely-used SMPL model~\cite{loper2015smpl} as our mesh representation, for its ability to express various human bodies using low dimensional parametric structures.

As mentioned before, this problem suffers from ambiguity issues because of the occlusions and the camera projection.
Directly training on one CNN as the regressor can easily lead to the model getting stuck in local minima, and it cannot be adapted to an arbitrary number of input images.
Inspired by the residual network structure~\cite{he2016deep}, we propose a multi-view multi-stage framework (Sec.~\ref{sec4}) to address this problem.
Since real-world datasets suffer from limited foreground/background textures and ground-truth pose and shape parameters, we make use of synthetic data as additional training samples (Sec.~\ref{sec5}) so that the model can be trained to be more shape-aware.

%% file: 4-network.tex
\vspace{-0.5em}
\section{Model Architecture}
\captionsetup{font=small}
\label{sec4}
\begin{figure*}
    \centering
  \includegraphics[height=2.4in]{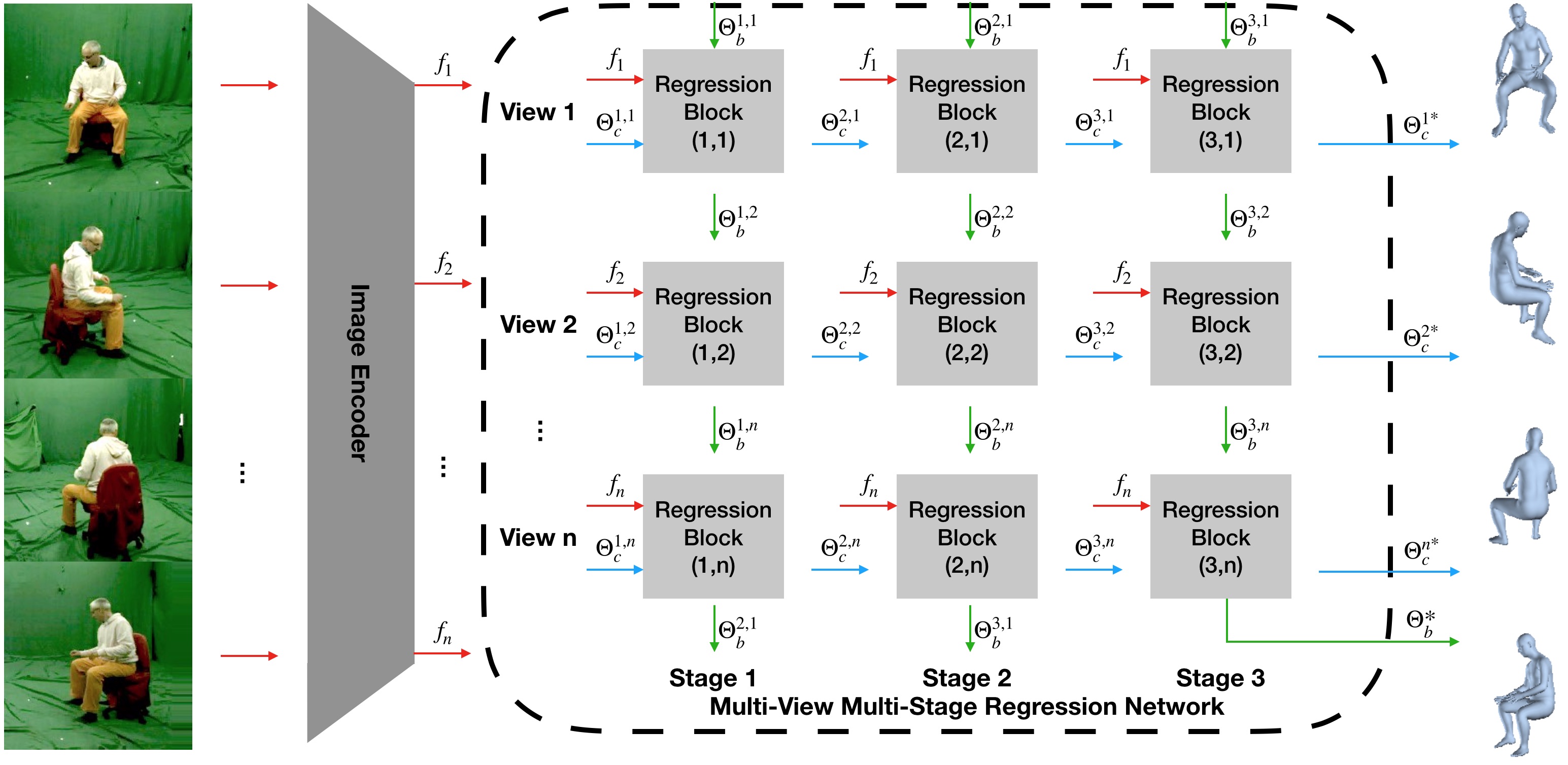}
  \caption{The network structure. Multi-view images are first passed through an image encoder to get feature vectors $f_1,...,f_n$. With initial guesses of the camera parameters $\Theta_c^{1,i}$ and the human body parameters $\Theta_b^{1,1}$, the network starts to estimate the parameters stage by stage and view by view. Each regression block at the $i^{th}$ stage and the ${j^{th}}$ view regresses the corrective values from image feature $f_j$ (red) and previous guesses $\Theta_c^{i,j}$ (blue) and $\Theta_b^{i,j}$ (green). The results will be added up to the input values and passed to future blocks. While the new human body parameters (green) can be passed to the next regression block, the view-specific camera parameters (blue) can only be passed to the next stage of the same view. Finally, the predictions of the $n$ views in the last stage are outputted to generate the prediction.}
\vspace{-1em}
  \label{fig4-1}
\end{figure*}
In this section, we describe the configuration of our network model.
As shown in Fig.~\ref{fig4-1}, we iteratively run our model for several stages of error correction.
Inside each stage, the multi-view image input is passed on one at a time.
At each step, the shared-parameter prediction block computes the correction based on the image feature and the input guesses.
We estimate the camera and the human body parameters at the same time, projecting the predicted 3D joints back to 2D for loss computation.
The estimated pose and shape parameters are shared among all views, while each view maintains its camera calibration and the global rotation.
The loss at each step is the sum of the joint loss and the human body parameter loss:
\begin{equation}
    L_i=\lambda_0 L_{2Djoint}+\lambda_1L_{3Djoint}+L_{SMPL}
\end{equation}
where $\lambda_0$ and $\lambda_1$ scale the units and control the importance of each term.
We use L1 loss on 2D joints and L2 loss on others. $L_{SMPL}$ is omitted if there is no ground-truth.

\vspace{-0em}
\subsection{3D Body Representation}
\label{sec4-1}
We use the Skinned Multi-Person Linear (SMPL) model~\cite{loper2015smpl} as our human body representation.
It is a generative model trained from human mesh data.
The pose parameters are the rotations of 23 joints inside the body, and the shape parameters are extracted from PCA.
Given the pose and shape parameter, the SMPL model can then generate a human body mesh consisting of 6980 vertices:
\begin{equation}
    \mathbf{X}(\theta,\beta)=\mathbf{W}\mathbf{G}(\theta) (\mathbf{X}_0+\mathbf{S}\beta+\mathbf{P} \mathbf{R}(\theta))
\end{equation}
where $\mathbf{X}\in \mathbb{R}^{6980}\times\mathbb{R}^3$ is the computed vertices, $\theta \in \mathbb{R}^{72}$ are the rotations of each joint plus the global rotation, $\beta \in \mathbb{R}^{10}$ are the PCA coefficients, $\mathbf{W}, \mathbf{S}$ and $\mathbf{P}$ are trained matrices, $\mathbf{G}(\theta)$ is the global transformation, $\mathbf{X_0}$ are the mean body vertices, and $\mathbf{R}(\theta)$ is the relative rotation matrix.

For the camera model, we use orthogonal projection since it has very few parameters and is a close approximation to real-world cameras when the subject is sufficiently far away, which is mostly the case.
We project the computed 3D body back to 2D for loss computation:
\begin{equation}
    \mathbf{x}=s\mathbf{X}(\theta,\beta)\mathbf{R}^T+\mathbf{t}
\end{equation}
where $\mathbf{R}\in\mathbb{R}^{2}\times\mathbb{R}^3$ is the orthogonal projection matrix, $s$ and $\mathbf{t}$ are the scale and the translation, respectively.

\subsection{Scalable Multi-View Framework}
\label{sec4-2}

Our proposed framework uses a recurrent structure, making it a universal model applicable to the input of any number of views.
At the same time, it couples the shareable information across different views so that the human body pose and shape can be optimized using image features from all views.
As shown in Fig.~\ref{fig4-1}, we use a multi-view multi-stage framework to couple multiple image inputs, with shared parameters across all regression blocks.
Since the information from multiple views can interact with each other multiple times, the regression needs to run for several iterative stages.
We choose to explicitly express this shared information as the predicted human body parameter since it is meaningful and also contains all of the information of the human body.
Therefore the input of a regression block is the corresponding image feature vector and the predicted camera and human body parameters from the previous block.
Inspired by the residual networks~\cite{he2016deep}, we predict the corrective values instead of the updated parameters at each regression block to prevent gradient vanishing.

We have $n$ blocks at each stage, where $n$ is the number of views.
Since all the input images contain the same human body with the same pose, these $n$ blocks should output the same human-specific parameters but possibly different camera matrices.
Thus we share the human parameter output across different views and the camera transformation across different stages of the same view.
More specifically, the regression block at the $i^{th}$ stage and the $j^{th}$ view takes an input of $(f_j,\Theta_c^{i,j},\Theta_b^{i,j})$, and outputs the correction $\Delta\Theta_c^{i,j},\Delta\Theta_b^{i,j}$, where $f_j$ denotes the $j^{th}$ image feature vector, $\Theta_c^{i,j}$ is the camera matrices and $\Theta_b^{i,j}$ is the human parameters.
After that, we pass $\Theta_c^{i+1,j}=\Theta_c^{i,j}+\Delta\Theta_c^{i,j}$ to the next \textbf{stage} of the block at the same view, while we pass $\Theta_b^{i,j+1}=\Theta_b^{i,j}+\Delta\Theta_b^{i,j}$ to the next \textbf{block} of the chain (Fig.~\ref{fig4-1}).
At last, we compute the total loss as the average of the prediction of all $n$ views in the final stage.
Different from static multi-view CNNs which have to fix the number of inputs, we make use of the RNN-like structure in a cyclic form to accept any number of views, and avoid the gradient vanishing by using the error correction framework.

\vspace{-0em}
\subsection{Training and Inferring}
\label{sec4-3}
Intuitively we use $n=4$ in our training process, since providing front, back, left, and right views can often give sufficient information about the human body.
We choose a random starting view from the input images to account for the potential correlation between the first view and the initial guess.
A specific order of the input views is not required since (a) the network parameters of each regression block are identical, and (b) none of the camera rotation information are shared among different views.
To make use of large public single-view datasets, we copy each instance to 4 identical images as our input.

During inference, our framework can adapt to images with any number of views $n$ as shown below.
If $n\leq 4$, we use the same structure as used for training.
We can pad any of the input images to fill up the remaining views.
As each view is independent in terms of global rotation, the choice of which view to pad does not matter.
If $n > 4$, we extend our network to $n$ views.
Since this is an error-correction structure, the exceeded values introduced by extra steps can be corrected back.
Note that the number of camera parameter corrections of each view always remains the same, which is the number of stages.

\begin{figure}
\centering
  \includegraphics[width=3in]{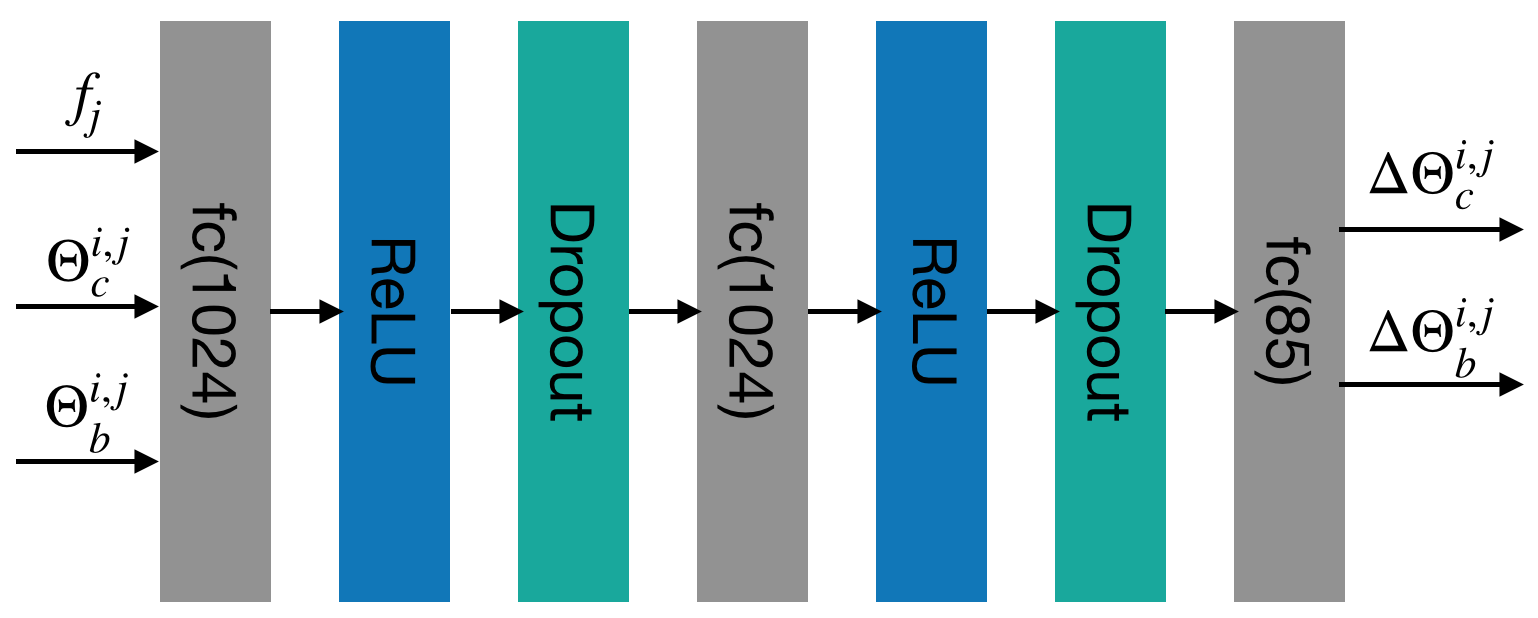}
  \vspace{-1em}
  \caption{Detailed network structure of the regression block at the $i^{th}$ stage and the $j^{th}$ view. $f_j$ denotes the image feature of the $j^{th}$ view, $\Theta_c^{i,j}$ denotes the camera parameters, and $\Theta_b^{i,j}$ denotes the human body parameters.}
  \label{fig4-2}
\vspace{-1.5em}
\end{figure}

\vspace{-0.5em}
\subsection{Implementation Details}
\label{sec4-4}
During training, besides our synthetic dataset for enhancing the shape estimation (detailed discussion in Sec.~\ref{sec5}), we train on MS-COCO~\cite{lin2014microsoft}, MPI\_INF\_3DHP~\cite{mono-3dhp2017} and Human3.6M~\cite{ionescu2014human3} datasets.
Each mini-batch consists of half single view and half multi-view samples.
Different from HMR~\cite{hmrKanazawa17}, we do not use the discriminator.
This is because (a) we initialized our parameters as the trained model of HMR~\cite{hmrKanazawa17}, (b) the ground-truth given by our dataset serves as the regularization to prevent unnatural pose not captured by joint positions (\eg foot orientations), and most importantly, (c) the ground-truth SMPL parameters from their training dataset does not have sufficient shape variety.
Enforcing the discriminator to mean-shape biased dataset will prevent the model to predict extreme shapes.
We use 50-layer ResNet-v2~\cite{he2016identity} for image feature extraction.
The detailed structure inside the regression block is shown in Fig.~\ref{fig4-2}.
We fix the number of stages as 3 throughout the entire training and all testing experiments.
The learning rate is set to $10^{-5}$, and the training lasts for 20 epochs.
Training on a GeForce GTX 1080 Ti GPU takes about one day.
Our synthetic dataset will be released with the paper.

%% file: 5-data.tex
\section{Data Preparation}
\label{sec5}
To the best of our knowledge, there is no public real-world dataset that captures motion sequences of human bodies, annotated with pose and shape (either using a parametric model or raw meshes), with considerable shape variation and loose garments.
This lack of data, in turn, forces most of the previous human body estimations to focus only on joints.
The most recent work~\cite{hmrKanazawa17} that recovers both pose and shape of human body does not impose an explicit shape-related loss function, so their model is not aware of varying human body shapes.
In order to make our model shape-aware under clothing, we need data with ground-truth human body shapes where the garments should be dressed rather than pasted on the skin.
A large amount of data is needed for training; sampling real-world data that captures the ground-truth shape parameters is both challenging and time-consuming.
We choose an alternate method --- using synthesized data. In this section, we propose an automatic pipeline to generate shape-aware training data, to enhance the shape estimation performance.

\vspace{-0em}
\subsection{Parameter Space Sampling}
We employ the SMPL model~\cite{loper2015smpl}, which contains pose and shape parameters for human body.
Pose parameters are rotation angles of joints.
To sample meaningful human motion sequences in daily life, we use the CMU MoCap dataset~\cite{cmumocap} as our pose subspace.
The shape parameters are principle component weights.
It is not ideal to sample the shape parameters using Gaussian distribution; otherwise there will be many more mean-shape values than extreme ones, resulting in an unbalanced training data.
To force the model to be more shape-aware, we choose to uniformly sample values at $[\mu-3\sigma,\mu+3\sigma]$ instead, where $\mu$ and $\sigma$ represent the mean value and standard deviation of the shape parameters.

\vspace{-0em}
\subsection{Human Body Motion Synthesis}
\label{sec5-2}
After combining CMU MoCap pose data with the sampled shape parameters, it is likely that the human mesh generated by the SMPL model has inter-penetration due to the shape difference.
Since inter-penetration is problematic for cloth simulation, we design an optimization scheme to avoid it in a geometric sense:
\begin{equation}
\begin{aligned}
\min \|\mathbf{x}-\mathbf{x_0}\|\quad s.t.\quad g(\mathbf{x})+\epsilon\leq 0
\end{aligned}
\end{equation}
where $\mathbf{x}$ and $\mathbf{x}_0$ stand for the vertex positions, $g(\mathbf{x})$ is the penetration depth, and $\epsilon$ is designed to reserve space for the garment.
The main idea here is to avoid inter-penetrations by popping vertices out of the body, but at the same time keeping the adjusted distance as small as possible, so that the body shape does not change much.
This practical method works sufficiently well in most of the cases.

\vspace{-0em}
\subsection{Cloth Registration and Simulation}
\label{sec5-3}
Before we can start to simulate the cloth on each body generated, we first need to register them to the initial pose of the body.
To account for the shape variance of different bodies, we first manually register the cloth to one of the body meshes.
We mark the relative rigid transformation $T$ of the cloth.
For other body meshes, we compute and apply the global transformation, including both the transformation $T$ and the scaling between two meshes.
At last, we use the similar optimization scheme described in Sec.~\ref{sec5-2} to avoid any remaining collisions since it can be assumed that the amount of penetration after the transformation is small.

We use ArcSim~\cite{narain2012adaptive} as the cloth simulator. 
We do not change the material parameters during the data generation.
However, we do randomly sample the tightness of the cloth.
We generally want both tight and loose garments in our training data.

\vspace{-0em}
\subsection{Multi-View Rendering}
\label{sec5-4}
\begin{figure}
    \centering
  \includegraphics[height=1.7in]{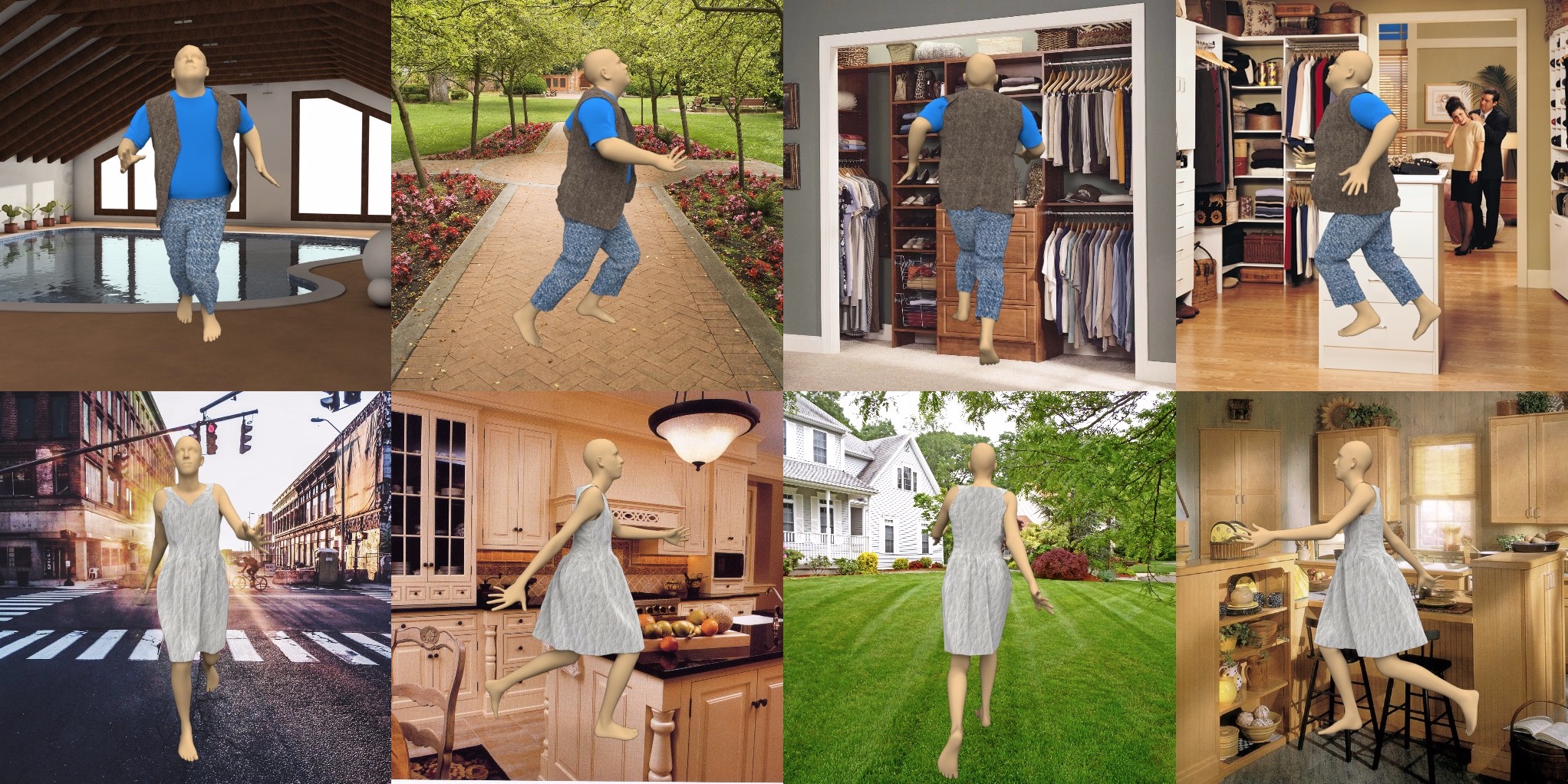}
  \vspace{-1.5em}
  \caption{Examples of rendered synthetic images. We use a large number of real-world backgrounds and cloth textures so that the rendered images are realistic and diverse.}
  \label{fig5-1}
\vspace{-1.5em}
\end{figure}
We randomly apply different background and cloth textures in different sets of images.
We keep the same cloth textures but apply different background across different views.
We use the four most common views (front, back, left, and right), which are defined \wrt the initial human body orientation and fixed during the rendering.
We sample 100 random shapes and randomly apply them to 5 pose sequences in the CMU MoCap dataset (slow and fast walking, running, dancing, and jumping).
After resolving collisions described in~\ref{sec5-3}, we register two sets of clothes on it, one with a dress and the other with a t-shirt, pants, and jacket (Fig.~\ref{fig5-1}).
The pose and garment variety is arguably sufficient because (a) they provide most commonly seen poses and occlusions, and (b) it is an auxiliary dataset providing shape ground-truth which is jointly trained with real-world datasets that have richer pose ground-truth.
We render two instances of each of the simulated frames, with randomly picked background and cloth textures.
Given an average of 80 frames per sequence, we have generated 32,000 instances, with a total number of 128,000 images.
We set the first 90 shapes as the training set and the last 10 as the test set.
We ensure the generalizability across pose and clothing by coupling our dataset with other datasets with joint annotations (Sec.~\ref{sec4-4}).

%% file: 6-result.tex
\vspace{-0.5em}
\section{Results}
\label{sec6}
We use the standard test set in Human3.6M and the validation set of MPI\_INF\_3DHP
to show the performance gain by introducing multi-view input.
Since no publicly available dataset has ground-truth shape parameters or mesh data, or data contains significantly different shapes from those within the normal range of BMI (\eg overweight or underweight bodies), we test our model against prior work (as the baseline) using the 
synthetic test set.
Also, we test on real-world images to show that our model is more {\em shape-aware} than the baseline method -- qualitatively using online images and quantitatively using photographs taken with hand-held cameras.

Our method does not assume prior knowledge of the camera calibration so the prediction may have a scale difference compared to the ground-truth.
There is also extra translation and rotation due to image cropping.
To make a fair comparison against other methods, we report the metrics after a rigid alignment, following~\cite{hmrKanazawa17}.
We also report the metrics before rigid alignment in the appendix.

\vspace{-0.5em}
\subsection{Ablation Study}
\label{sec6-1}
We conduct an ablation study to show the effectiveness of our model and the synthetic dataset.
In the experiments, HMR~\cite{hmrKanazawa17} is fine-tuned with the same learning setting.

\vspace{-1em}
\subsubsection{Pose Estimation}
We tested our model on datasets using multi-view images to demonstrate the strength of our framework.
We use \emph{Mean Per Joint Position Error} (MPJPE) of the 14 joints of the body, as well as \emph{Percentage of Correct Keypoints} (PCK) at the threshold of 150mm along with \emph{Area Under the Curve} (AUC) with threshold range 0-150mm~\cite{mehta2017monocular} as our metrics.
PCK gives the fraction of keypoints within an error threshold, while AUC computes the area under the PCK curve, presenting a more detailed accuracy within the threshold.

We use the validation set of MPI\_INF\_3DHP~\cite{hmrKanazawa17} as an additional test dataset since it provides multi-view input.
It is not used for validation during our training.
We also evaluated the original test set, which consists of single-view images.
Please refer to our appendix in the supplementary document for this comparison result.

\noindent
{\bf Comparison: } As shown in Table~\ref{tab6-1} and~\ref{tab6-2}, under the same training condition, our model in single-view has similar, if not better, results in all experiments.
Meanwhile, our model in multi-view achieves much higher accuracy.

\begin{table}[h]
\captionsetup{font=small}
\begin{center}
\begin{tabular}{c|c|c}
\hline
Method & \begin{tabular}{@{}c@{}}MPJPE\\w/ syn. training\end{tabular} & \begin{tabular}{@{}c@{}}MPJPE\\w/o syn. training\end{tabular}\\
\hline
HMR & 60.14 &58.1 \\
Ours (single)&58.55 &	59.09 \\
Ours (multi)  & \textbf{45.13} &	\textbf{44.4} \\
\hline
\end{tabular}
\end{center}
\vspace{-1.5em}
\caption{Comparison results on Human3.6M using MPJPE. Smaller errors implies higher
accuracy.}
\label{tab6-1}
\vspace{-1em}
\end{table}

\begin{table}[h]
\begin{center}
\begin{tabular}{c|c|c}
\hline
Method & \begin{tabular}{@{}c@{}}PCK/AUC/MPJPE\\w/ syn. training\end{tabular} & \begin{tabular}{@{}c@{}}PCK/AUC/MPJPE\\w/o syn. training\end{tabular}\\
\hline
HMR & 86/49/89 &	88/52/83 \\
Ours (single)& 88/52/84 &	87/52/85 \\
Ours (multi)  & \textbf{95}/\textbf{63}/\textbf{62} &	\textbf{95}/\textbf{65}/\textbf{59} \\
\hline
\end{tabular}
\end{center}
\vspace{-1.5em}
\caption{Comparison results on MPI\_INF\_3DHP in PCK/AUC/ MPJPE. Better results have higher PCK/AUC and lower MPJPE.}
\label{tab6-2}
\vspace{-2em}
\end{table}

\vspace{-0em}
\subsubsection{Shape Estimation}
\label{sec6-2}
To the best of our knowledge, there is no publicly available dataset that provides images with the captured human body mesh or other representation among a sufficiently diverse set of human shapes.
Since most of the images-based datasets are designed for joint estimation, we decide to use our synthetic test dataset for large-scale statistical evaluation, and later compare with~\cite{hmrKanazawa17} using real-world images.

Other than MPJPE for joint accuracy, we use the Hausdorff distance between two meshes to capture the shape difference to the ground-truth.
The Hausdorff distance is the maximum shortest distance of any point in a set to the other set, defined as follows:
\begin{equation}
    d(V_1,V_2)=\max(\hat{d}(V_1,V_2),\hat{d}(V_2,V_1))
\end{equation}
\begin{equation}
    \hat{d}(V_1,V_2)=\max_{\mathbf{u}\in V_1}\min_{\mathbf{v}\in V_2}\|\mathbf{u}-\mathbf{v}\|^2
\end{equation}
where $V_1$ and $V_2$ are the vertex set of two meshes in the same ground-truth pose, in order to negate the impact of different poses.
Intuitively a Hausdorff distance of $d$ means that by moving each vertex of one mesh by no more than $d$ away, two meshes will be exactly the same.

\begin{table}[h]
\begin{center}
\begin{tabular}{c|c|c}
\hline
Method & \begin{tabular}{@{}c@{}}MPJPE/HD\\w/ syn. training\end{tabular} & \begin{tabular}{@{}c@{}}MPJPE/HD\\w/o syn. training\end{tabular}\\
\hline
HMR & 42/83 &	89/208 \\
Ours (single)& 44/65 & 	102/283 \\
Ours (multi)  & \textbf{27}/\textbf{53} &	\textbf{84}/273 \\
\hline
\end{tabular}
\vspace{-1.5em}
\end{center}
\caption{Comparison results on our synthetic dataset in MPJPE/Hausdorff Distance(HD). Better results have lower values.} 
\label{tab6-3}
\vspace{-1.5em}
\end{table}

As shown in Table~\ref{tab6-3}, our model with multi-view input achieves the smallest error values, when compared to two other baselines.
After joint-training with synthetic data, all models perform better in shape estimation, while maintaining similar results using other metrics (Table~\ref{tab6-1} and~\ref{tab6-2}), i.e. they do not overfit.
The joint errors of the HMR~\cite{hmrKanazawa17} are fairly good, so they can still recognize the synthesized human in the image.
However, a larger Hausdorff distance indicates that they lose precision on the shape recovery.

Adding our synthetic datasets for training can effectively address this issue and thereby provide better shape estimation.
We achieved a much smaller Hausdorff distance (with syn. training) even only using single view.
This is because our refinement framework is effectively deeper, aiming at not only the pose but also the shape estimation, which is much more challenging than the pose-only estimation.
With the same method, multi-view inputs can further improve the accuracy of shape recovery compared to results using only one single-view image.

\subsection{Comparisons with Multi-View Methods}
\label{sec6-2}
Since other multi-view methods only estimate human poses but not the entire body mesh, we compare the pose estimation results to them in Human3.6M.
As shown in Table~\ref{tab6-4}, we achieved state-of-the-art performance even when camera calibration is unknown and no temporal information is provided.
As stated in Sec.~\ref{sec6}, unknown camera parameters result in a scaling difference to the ground-truth, so the joint error would be worse than what it actually is.
After the Procrustes alignment that accounts for this effect, our method achieves the best MPJPE compared to other methods.
Another potential source of the error is that our solution is constrained in a parametric subspace, while other methods output joint positions directly.
In contrast, our method computes the entire human mesh in addition to joints and the result can be articulated and animated directly.
\begin{table*}
\begin{center}
\begin{tabular}{c|c|c|c|c|c|c}
\hline
Method & MPJPE & Known Camera?&Run Time&Temporal Opt?&Articulated?&Shape?\\
\hline
Rhodin \etal~\cite{rhodin2016general}&-&Yes&0.025fps&Yes&No&Mix-Gaussian\\
Rhodin \etal~\cite{rhodin2018unsupervised}&98.2&Yes&-&Yes&No&No\\
Pavlakos \etal~\cite{pavlakos2017harvesting}&56.89&Yes&-&No&No&No\\
Trumble \etal~\cite{trumble2017total}&87.3&Yes&25fps&Yes&No&No\\
Trumble \etal~\cite{trumble2018deep}&62.5&Yes&3.19fps&Yes&No&Volumetric\\
N{\'u}{\~n}ez \etal~\cite{nunez2019multiview}&54.21&Yes&8.33fps&Yes&No&No\\
Tome \etal~\cite{tome2018rethinking}&52.8&Yes&-&No&No&No\\
\hline
Ours & 79.85 &\multirow{2}{*}{No}&\multirow{2}{*}{33fps}&\multirow{2}{*}{No}&\multirow{2}{*}{Yes}&\multirow{2}{*}{Parametric}\\
Ours (PA) & \textbf{45.13}&&&&\\
\hline
\end{tabular}
\vspace{-1.5em}
\end{center}
\caption{Comparison on Human3.6M with other multi-view methods. Our method has comparable performance with previous work even without the assistance of camera calibration or temporal information. PA stands for Procrustes Aligned results for ours.}
\label{tab6-4}
\vspace{-1em}
\end{table*}

\vspace{-0.5em}
\subsection{Real-World Evaluations}

\begin{table}
\begin{center}
\begin{tabular}{c|c|c}
\hline
Method & Standing & Sitting\\
\hline
HMR~\cite{hmrKanazawa17} & 7.72\% &	7.29\% \\
BodyNet~\cite{varol2018bodynet}& 13.72\% & 	29.30\% \\
Ours (single)  & 6.58\% &	10.18\% \\
Ours (multi) & \textbf{6.23\%} &	\textbf{5.26\%} \\
\hline
\end{tabular}
\end{center}
\vspace{-1.5em}
\caption{Comparison results on tape-measured data using average relative errors (lower the better).}
\label{tab6-5}
\vspace{-1.5em}
\end{table}
We first conduct a study on how our method performs differently with either single- or multi-view inputs under various conditions.
Our test subjects have two poses: standing and sitting, and the model is additionally tested on two sets of variants from the images.
One is slightly dimmed, and the other has a large black occlusion at the center of the first image.
We use the percentage of errors from common body measurements used by tailors (\ie lengths of neck, arm, leg, chest, waist, and hip), which is obtained using direct tape measurements on the subjects.
We report the average relative error in Table~\ref{tab6-5}.
The detailed errors of each measurement are also provided in the appendix.
It is observed that single-view results are affected by the ``occluded sitting'' case, while the multi-view input can largely reduce the error.
The reason why HMR is not impacted is that they uniformly output average human shapes for all input images.
We also report results from BodyNet~\cite{varol2018bodynet}.
BodyNet outputs voxelized mesh and needs a time-consuming optimization to output the SMPL parameters.
Its accuracy largely depends on the initial guess.
Therefore, it resulted in a large amount of errors on the ``sitting'' case.

\begin{figure}
    \centering
    \begin{subfigure}[t]{0.145\textwidth}
    \centering
  \includegraphics[height=1.6in]{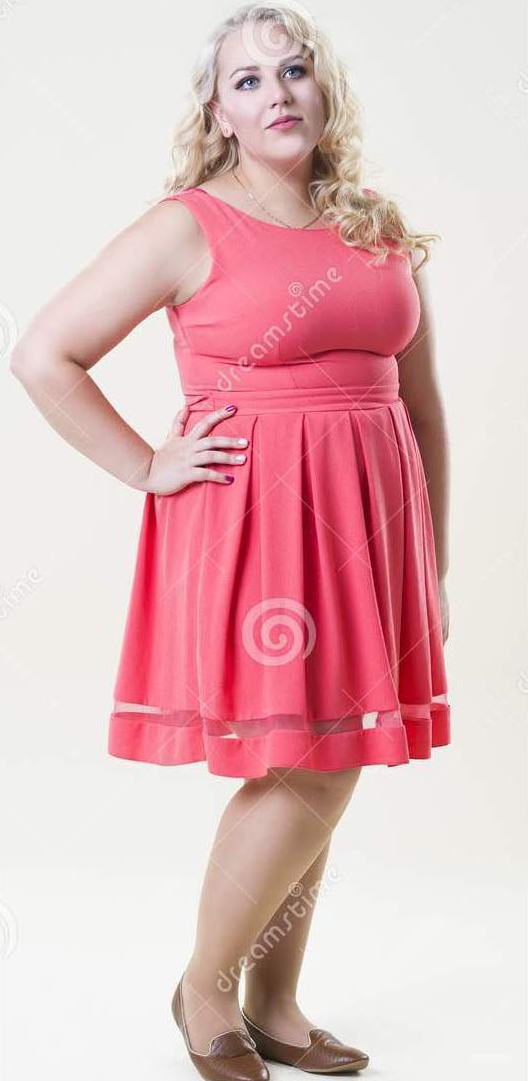}
  \caption{The input image.}
    \end{subfigure}
    ~
    \begin{subfigure}[t]{0.145\textwidth}
    \centering
  \includegraphics[height=1.6in]{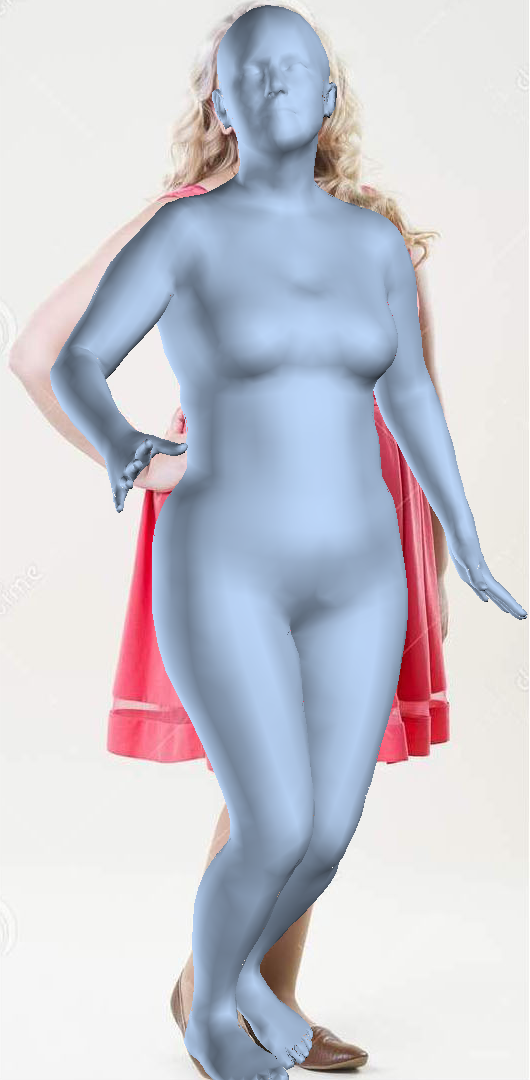}
  \caption{Our result.}
    \end{subfigure}
    ~
    \begin{subfigure}[t]{0.145\textwidth}
    \centering
  \includegraphics[height=1.6in]{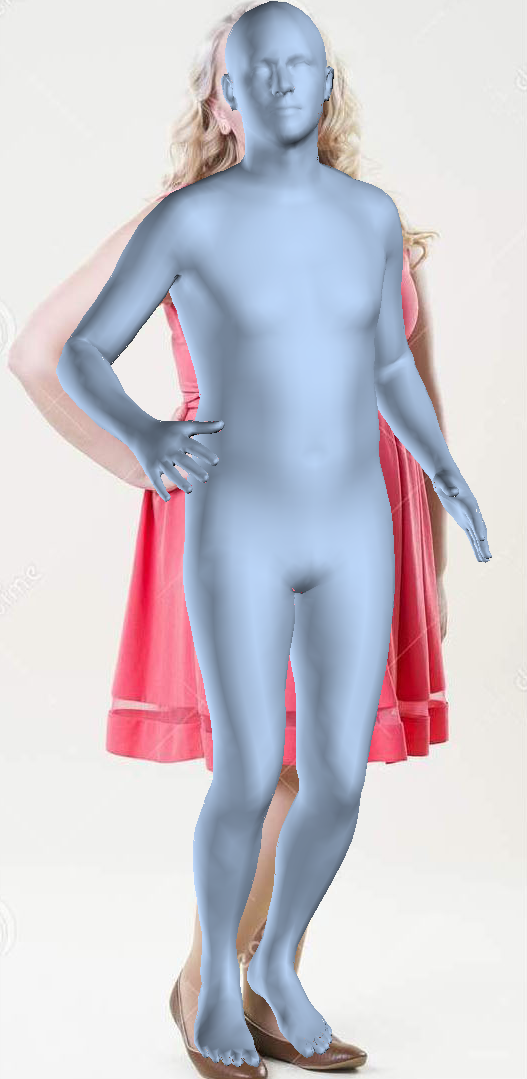}
  \caption{HMR.}
    \end{subfigure}
  \vspace*{-0.5em}
  \caption{Prediction results compared to HMR. Our model can better capture the shape of the human body. The recovered legs and chest are closer to the person in the image.}
  \label{fig6-2}
\vspace{-2em}
\end{figure}

We also tested our model on other online images, where no such measurement can be done.
As shown in Fig.~\ref{fig6-2}, HMR~\cite{hmrKanazawa17} can predict the body pose but fails on inferring the person's shape.
On the contrary, our model not only refines the relative leg orientations but also largely respects and recovers the original shape of the body.
More examples are shown in our supplemental document and video.

\vspace{-0.5em}
\subsection{Multi-View Input in Daily Life}
\label{sec6-3}
It is often difficult to have multiple cameras from different view angles capturing a subject simultaneously.
Our model has the added benefit of not requiring the multi-view input be taken with the exact same pose.
As the model has an error correction structure, it can be applied as long as the poses of the four views are not significantly different.
We do not impose any assumptions on the background, so the images can be even taken with a fixed camera and a ``rotating'' human subject, which is the typically case when the method is used in applications like virtual try-on.

%% file: 7-conclusion.tex
\vspace{-0.5em}
\section{Conclusion and Future Work}
\vspace*{-0.5em}
\label{sec7}
We proposed a novel multi-view multi-stage framework for pose and shape estimation.
The framework is trained on datasets with at most 4 views but can be naturally extended to an arbitrary number of views. 
Moreover, we introduced a physically-based synthetic data generation pipeline to enrich the training data, which is very helpful for shape estimation and regularization of end effectors that traditional datasets do not capture.
Experiments have shown that our trained model can provide equally good pose estimation as state-of-the-art using single-view images, while providing considerable improvement on pose estimation using multi-view inputs and a better shape estimation across all datasets.

While synthetic data improves the diversity of human bodies with ground-truth parameters, a more convenient cloth design and registration are needed to minimize the performance gap between real-world images and synthetic data.
In addition, other variables such as hair, skin color, and 3D backgrounds are subtle elements that can influence the perceived realism of the synthetic data at the higher expense of a more complex data generation pipeline.
With the recent progress in image style transfer using GAN~\cite{mueller2018ganerated}, a promising direction is to transfer the synthetic result to more realistic images to further improve the learning result.

\vspace*{0.5em}
\noindent
{\bf Acknowledgement:} This work is supported by National Science Foundation and Elizabeth S. Iribe Professorship.

%% file: appendix.tex
\clearpage
\begin{appendices}

\section{Extra Test Results}
Table~\ref{taba-1} and~\ref{taba-3} shows the test results before Procrustes Alignment in MPI\_INF\_3DHP validation set and Human3.6M, respectively.
The same conclusion about over-fitting and multi-view improvement as the main text can also be drawn from these data.

Table~\ref{taba-2} shows the result in MPI\_INF\_3DHP test dataset.
Since there is only one view fed into the model, the results are similar.

    \begin{figure*}[!hb]
    \centering
  \includegraphics[width=\textwidth]{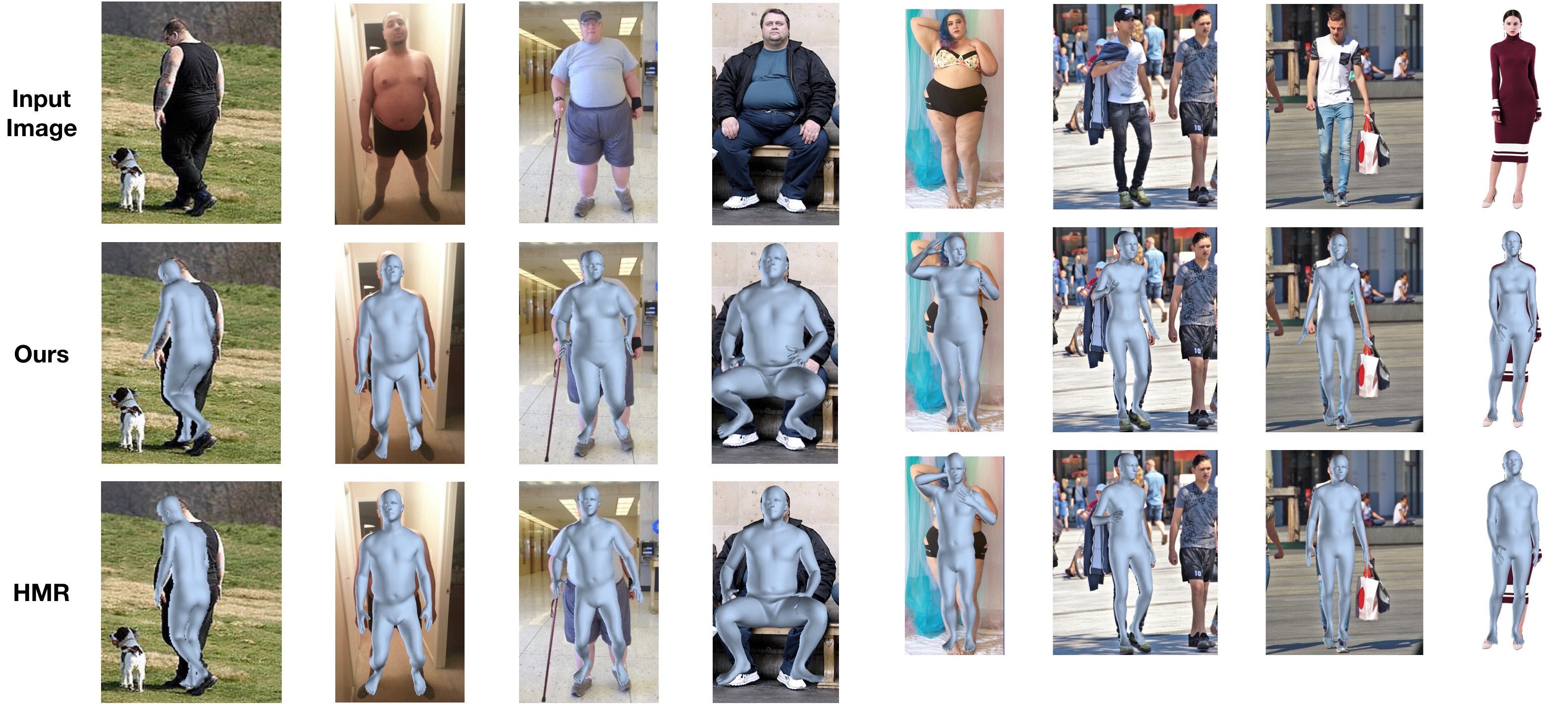}
  \caption{Results on images with varying pose and shape. The top row is the input image. The middle row shows our recovery results, and the bottom row shows the results from HMR~\cite{hmrKanazawa17}. Ours achieves better shape recovery results.}
  \label{figa-2}
  \vspace{-1em}
    \end{figure*}
    \begin{figure*}[!hb]
    \centering
  \includegraphics[width=0.7\textwidth]{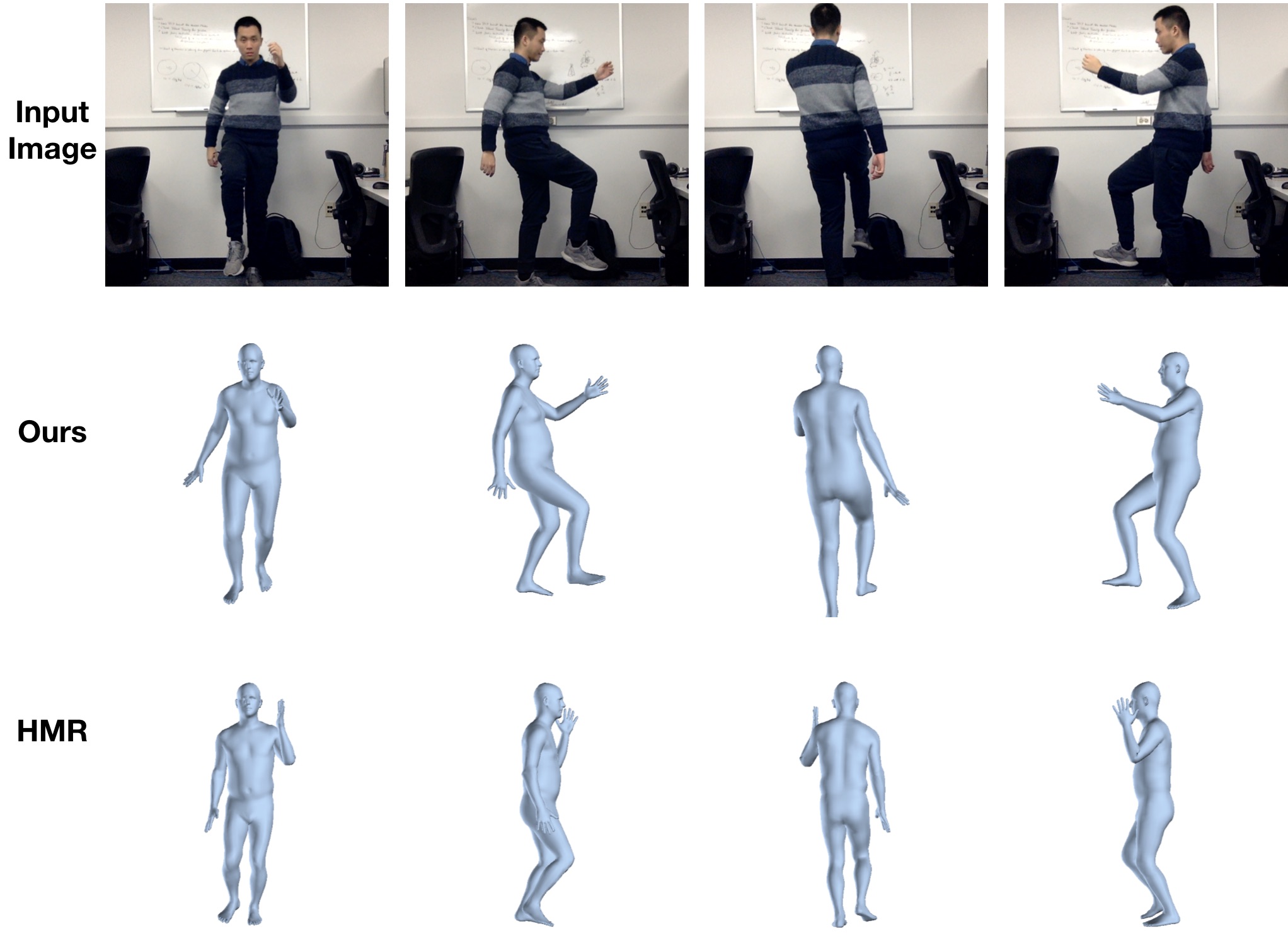}
  \caption{Results on real-world multi-view images. The top row is the input image. The middle row shows our recovery results, and the bottom row shows the results from HMR~\cite{hmrKanazawa17}. HMR is only given the front view as input. Ours achieves better pose recovery results due to more view angles.}
  \label{figa-3}
  \vspace{-1em}
    \end{figure*}

\begin{table}[h]
\begin{center}
\begin{tabular}{c|c|c}
\hline
Method & \begin{tabular}{@{}c@{}}PCK/AUC/MPJPE\\w/ syn. training\end{tabular} & \begin{tabular}{@{}c@{}}PCK/AUC/MPJPE\\w/o syn. training\end{tabular} \\
\hline
HMR~\cite{hmrKanazawa17} &66/33/141&71/36/129 \\
Ours (single) & 69/32/139&68/33/138 \\
Ours (multi) & 72/34/128&72/35/126\\
\hline
\end{tabular}
\end{center}
  \vspace{-1em}
\caption{Results on MPI\_INF\_3DHP, validation set, before Procrustes aligment.}
\label{taba-1}
\end{table}

\begin{table}[h]
\begin{center}
\begin{tabular}{c|c|c}
\hline
Method & \begin{tabular}{@{}c@{}}PCK/AUC/MPJPE\\w/ syn. training\end{tabular} & \begin{tabular}{@{}c@{}}PCK/AUC/MPJPE\\w/o syn. training\end{tabular} \\
\hline
HMR~\cite{hmrKanazawa17} &65/30/139&65/29/137 \\
HMR (PA) &84/47/91&85/48/89 \\
Ours & 65/29/142&66/29/137 \\
Ours (PA) & 85/49/89&86/49/89\\
\hline
\end{tabular}
\end{center}
  \vspace{-1em}
\caption{Results on MPI\_INF\_3DHP, test set. The results of~\cite{hmrKanazawa17} are tested on cropped images by Mask-RCNN~\cite{DBLP:journals/corr/HeGDG17} so the values have minor difference than their reported ones. Only single view is available in this dataset.}
\label{taba-2}
\end{table}
    
\section{Additional Results on Real-World Images}
As shown in Fig.~\ref{figa-2}, given similar joint estimation results, our model captures more image features that indicate the shape of the human body and thereby gives much better results in terms of human shape.
We can distinguish between fat (Column 1-5) and slim (Column 6-8) persons, and between male and female.
On the other hand, the output shapes from HMR are almost the same, which is around the mean shape value.
By incorporating the shape-aware synthetic dataset, our method largely improves the recovery when the input human body does not have an average shape.
We also tested with real-world multi-view images vs. single-view HMR.
We feed the front view of the subject to HMR but input all views into our model.
As shown in Fig.~\ref{figa-3}, the front view does not provide complete information of the subject pose, resulting in large pose errors on the limbs.
By sharing information from more views (most importantly side views in this case), our model can effectively reduce the ambiguity from the camera projection and thereby provide good pose estimations across all views.

\section{Comparison on Human3.6M with Single-View Methods}
\label{secc}
Table~\ref{taba-3} shows the comparison with single-view results.
As mentioned in the main text, the reason we don't have much better accuracy before rigid alignment is that:
\begin{itemize}
    \item Our method does not assume known camera, resulting in an unknown scaling difference to the real-world coordinates. After the Procrustes alignment, we achieved similar (and better with multi-view) performance.
    \item Our solution is constrained in a subspace.
Other methods output joint positions directly so they have more DOF and can be more accurate.
However, our output is more comprehensive, as it contains the entire human mesh in addition to joints and the result can be articulated and animated directly.
\end{itemize}

Compared to Kolotouros \etal~\cite{kolotouros2019convolutional}, our model is trained on a much more diverse dataset (\eg MS-COCO), which means that the accuracy may not be minimized on the specific subset (Human 3.6M).

\begin{table}
\begin{center}
\begin{tabular}{c|c|c}
\hline
Method & MPJPE & PA-MPJPE\\
\hline
Tome \etal~\cite{tome2017lifting}&88.39& -\\
Rogez \etal~\cite{rogez2017lcr}&87.7&71.6\\
Mehta \etal~\cite{mehta2017vnect}&80.5& -\\
Pavlakos \etal~\cite{pavlakos2017coarse}&71.9&\textbf{51.23}\\
Mehta \etal~\cite{mehta2017monocular}&68.6& -\\
Sun \etal~\cite{sun2017compositional}&\textbf{59.1}& -\\
Zhou \etal~\cite{zhou2016deep} & 107.26 & -\\
Debra \etal~\cite{dabral2018learning} & 55.5 & -\\
\hline
*Kolotouros \etal~\cite{kolotouros2019convolutional} & \textbf{74.7} & 51.9 \\
*Omran \etal~\cite{omran2018neural} & - & 59.9 \\
*Pavlakos \etal~\cite{pavlakos2018learning} & - & 75.9 \\
*HMR~\cite{hmrKanazawa17} & 87.97 & 58.1 \\
*Ours (single-view)& 88.34 & 58.55\\
*Ours (multi-view)  & 79.85 & \textbf{45.13}\\
\hline
\end{tabular}
\vspace{-1em}
\end{center}
\caption{Results on Human3.6M. Our method results in smaller reconstruction errors compared to HMR~\cite{hmrKanazawa17}. * indicates methods that output both 3D joints {\em and} shapes.}
\label{taba-3}
\vspace{-0.5em}
\end{table}

\section{Results Without Training on Synthetic Data}
\begin{figure}
    \centering
  \includegraphics[height=1.6in]{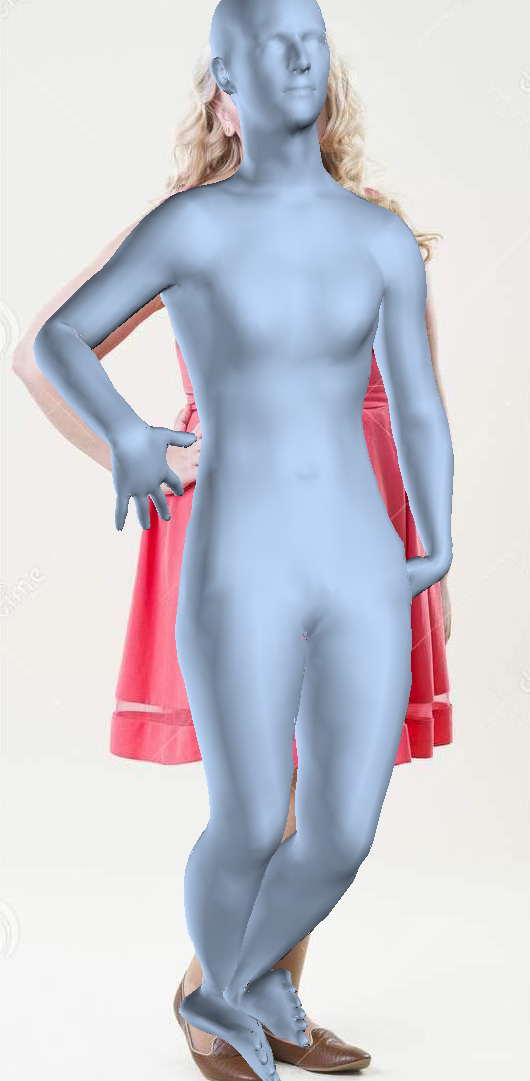}
  \caption{Our model trained without synthetic data.}
  \label{figa-1}
  \vspace{-1em}
    \end{figure}

We further tested another variant of our model, which is trained without synthetic data (Fig.~\ref{figa-1}).
It achieves better joint estimation, but the recovered human body does not seem to be visually correct, especially at the end-effectors.
This is because the joint-only supervision does not impose any constraints on the orientations of the end-effectors, resulting in an arbitrary guess.
The HMR model~\cite{hmrKanazawa17} avoids this by adding a discriminator, which however could have negative impact on shape estimations, as discussed in Sec.~\ref{sec4-4}.
Our synthetic dataset provides a supervision to not only the joint positions but also the rotations, hence the model will learn a prior at the end-effectors, demonstrating more natural results.

\section{Detailed Errors on Real World Evaluation}
The error percentages of each measure are shown in Table~\ref{taba-4}.
Since the length of the arm and leg can be seen clearly in the front view, both inputs provide a reasonably good estimation.
However, given more views, our model can significantly reduce the error on other measurements, especially on those of chest, waist, and hip.
We found that image illuminance has a negligible effect on the recovery result, which is due to the translation invariance of the convolutional layers.
Occlusion has a notable impact on the recovery using only a single-view image, given only one view of the human body.
However, by incorporating more views using our network model, the estimation can be considerably improved, indicating that the model using multi-view images is more robust to occlusion than with a single-view image as input.

\begin{table*}
    \centering
    \resizebox{2.1\columnwidth}{!}{
    \begin{tabular}{c|c c|c c|c c|c c|c c|c c}
         error \% & \multicolumn{4}{c|}{Regular} & \multicolumn{4}{c|}{Dimmed} & \multicolumn{4}{c}{Partly Occluded} \\ \hline
         input & \multicolumn{2}{c|}{Standing} & \multicolumn{2}{c|}{Sitting} & \multicolumn{2}{c|}{Standing} & \multicolumn{2}{c|}{Sitting} & \multicolumn{2}{c|}{Standing} & \multicolumn{2}{c}{Sitting}\\\hline
         \# of views &Single&Multi&Single&Multi&Single&Multi&Single&Multi&Single&Multi&Single&Multi\\\hline
         neck&1.12&	12.19&	0.048&	3.53&		0.58&	11.31&	0.39&	2.55&		0.45&	11.28&	22.11&	6.11\\\hline
         arm&4.76&	4.22&	8.03&	7.33&		6.21&	4.95&	8.10&	6.89&		5.20&	3.82&	7.20&	6.70\\\hline
         leg&6.65&	4.66&	2.94&	3.46&		5.18&	3.92&	2.83&	3.64&		2.53&	3.54&	4.94&	4.24\\\hline
         chest&4.59&	7.72&	8.40&	3.1&		6.20&	7.20&	8.13&	3.19&		19.80&	1.57&	30.04&	13.72\\\hline
         waist&2.42&	12.80&	5.46&	0.70&		3.73&	11.98&	5.01&	0.0084&		13.78&	8.52&	30.05&	10.61\\\hline
         hip&8.88&	0.62&	11.88&	5.83&		11.36&	0.12&	11.78&	5.50&		15.08&	1.65&	15.95&	7.54\\\hline
         error \% & \multicolumn{4}{c|}{Regular} & \multicolumn{4}{c|}{Dimmed} & \multicolumn{4}{c}{Partly Occluded} \\ \hline
         input & \multicolumn{2}{c|}{Standing} & \multicolumn{2}{c|}{Sitting} & \multicolumn{2}{c|}{Standing} & \multicolumn{2}{c|}{Sitting} & \multicolumn{2}{c|}{Standing} & \multicolumn{2}{c}{Sitting}\\\hline
         method &HMR&BodyNet&HMR&BodyNet&HMR&BodyNet&HMR&BodyNet&HMR&BodyNet&HMR&BodyNet\\\hline
         neck&10.4&	2.9&	4.8&	26.3&		8.4&	1.6&	4.6&	26.2&		9.2&	3.9&	5.7&	6.8\\\hline
         arm&6.1&	21.3&	9.8&	25.6&		8.6&	22.8&	9.7&	23.6&		8.1&	19.5&	9.7&	9.6\\\hline
         leg&7.9&	6.3&	1.8&	4.4&		4.3&	6.6&	1.8&	3.3&		5.1&	6.2&	2.1&	3.0\\\hline
         chest&11.2&	26.3&	11.7&	51.9&		11.7&	24.9&	11.6&	41.3&		11.9&	24.9&	11.6&	21.3\\\hline
         waist&9.4&	9.0&	8.7&	42.7&		9.4&	7.7&	8.5&	33.7&		9.7&	8.3&	8.4&	11.4\\\hline
         hip&1.25&	19.2&	7.8&	79.8&		3.5&	18.8&	7.7&	80&		2.9&	17&	5.5&	36.9\\
    \end{tabular}
    }
    \caption{Percentages of errors in common measurements of the human body under various lighting conditions using single-view vs. multi-view images. The multi-view model performs significantly better in estimating measurements of chest, waist, and hip, and is more robust, given variations in lighting and partial occlusion.}
    \label{taba-4}
\vspace{-1em}
\end{table*}

\section{Evaluation on \textit{3D People in the Wild}.}
We have conducted the evaluation on  \textit{3D People in the Wild} dataset.
As shown in Table~\ref{tab:3dpw}, although the dataset consists of single view images of only a few subjects with nearly standard shapes, our model achieved better accuracy over HMR, {while Alldieck \etal did not generalize well}.
The metric we used is mean joint error for pose, and mean vertex error with ground-truth pose for shape.

\section{Running Time}
The previous work~\cite{hmrKanazawa17} trained 55 epochs for 5 days, while ours trained 20 epochs for 1 day.
We list the training time here for reference, but it is actually not comparable since the batch size, epoch size and GPU type are not the same.
In our environment, the inference time of HMR~\cite{hmrKanazawa17} is 2 microseconds while ours takes 7.5 (per view).
This is because our network has a deeper structure to account for multiple views.

\begin{table}[]
\begin{small}
    \vspace{-0.25em}
    \centering
    \resizebox{\columnwidth}{!}{
    \begin{tabular}{c|c|c}
        Method & Mean Joint Err. & Mean Vertex Err. (GT Pose) \\\hline
        HMR & 93.77 & 21.71 \\
        Alldieck \etal~\cite{alldieck2019learning} & 169.61 & 47.07 \\
        Ours & 96.86 & 20.96 \\
    \end{tabular}
    }
    \caption{Evaluation on an unseen single-view dataset: \textit{3D People in the Wild.} Values are mean joint error for pose and mean vertex error with ground-truth pose. We have smaller error than Alldieck \etal.}
    \label{tab:3dpw}
    \vspace{-1.5em}
\end{small}
\end{table}

\end{appendices}